\documentclass{article}
\usepackage{spconf,amsmath,graphicx}
\usepackage{subfigure}
\usepackage{booktabs}
\usepackage{comment}
\usepackage{arydshln} 
\usepackage{multirow}
\usepackage{microtype}
\usepackage{color}
\usepackage{url}

\title{Building Markovian Generative Architectures over Pretrained LM Backbones for Efficient Task-Oriented Dialog Systems}
\name{Hong Liu$^{1,3}$, Yucheng Cai$^{1,3}$, Zhijian Ou$^{*,1,3}$\thanks{$^{*}$ Corresponding author (ozj@tsinghua.edu.cn).}, Yi Huang$^{2,3}$, Junlan Feng$^{2,3}$
}
\address{
$^{1}$Speech Processing and Machine Intelligence (SPMI) Lab, Tsinghua University, Beijing, China \\
  $^{2}$China Mobile Research Institute, Beijing, China \\
  $^{3}$Tsinghua University-China Mobile Communications Group Co., Ltd. Joint Institute, Beijing, China
}

\copyrightnotice{978-1-6654-7189-3/22/\$31.00~\copyright2023 IEEE}

\begin{document}
%
\maketitle
\begin{abstract}
Recently, Transformer based pretrained language models (PLMs), such as GPT2 and T5, have been leveraged to build generative task-oriented dialog (TOD) systems. A drawback of existing PLM-based models is their non-Markov architectures across turns, i.e., the whole history is used as the conditioning input at each turn. First, this brings inefficiencies in memory and computation. Furthermore, using the whole history increases model complexity and may hurt the training efficiency, especially when facing small amounts of labeled training data (the low-resource setting).
In this paper, motivated by the observation that dialog states could be viewed as Markov states, we propose to build Markovian Generative Architectures (MGA) over PLM backbones for efficient TOD systems.
Experiments on MultiWOZ2.1 show that in the rich-resource setting, the proposed Markov models reduce memory and time costs without performance degradation; in the low-resource setting, the training efficiency of the Markov models is more significant.
\end{abstract}
%
\section{Introduction}
\label{sec:intro}
\begin{figure}[t]
\centering
\includegraphics[width=0.9\linewidth]{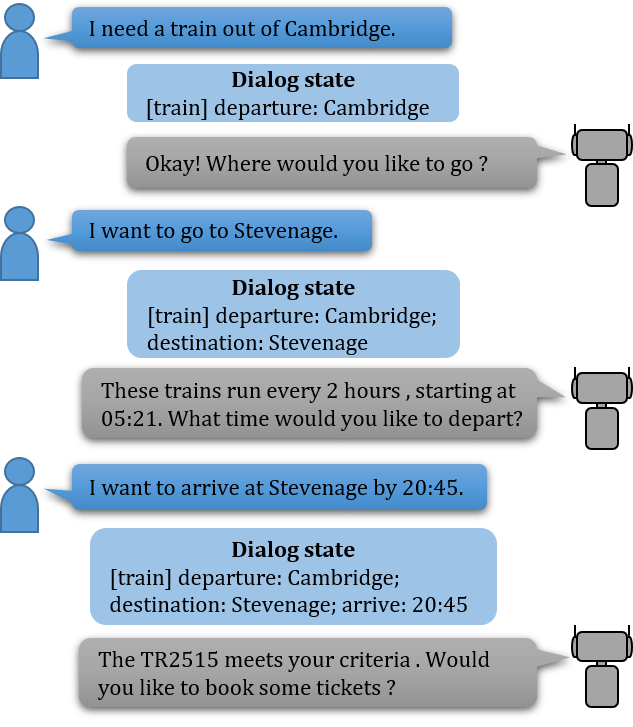}
\vspace{-1em}
\caption{An example of a task oriented dialog. It illustrates the accumulative nature of dialog state, which is usually defined to be a compact summary of dialog history and represented by a set of slot-value pairs.}
\label{fig:TOD}
\vspace{-1.5em}
\end{figure}

\vspace{-0.5em}
Task-oriented dialog (TOD) systems interact with users in natural languages through multiple turns to accomplish tasks.
At each turn, the system needs to parse the user utterance and track the dialog state, which is usually defined to be a compact summary of dialog history, as shown in Figure~\ref{fig:TOD}.
The dialog state is often represented by a set of slot-value pairs that determine the user's requirement.
Based on the tracked dialog state, the system will query a task-related database (DB), decide an action and generate a response.
The methodology for building TOD systems is gradually advancing from separate training of individual modules \cite{williams2016dialog,mrkvsic2017neural, dai2018tracking} to the end-to-end (E2E) trainable approach \cite{wen2017a, liu2017end, lei2018sequicity, fsdm, zhang2020task,gao2020paraphrase,zhang-etal-2020-probabilistic}.
In early E2E methods, the sequential turns of a dialog are usually modeled as a Markov process and realized over LSTM-based backbones.

Recently, self-attention based Transformer neural networks \cite{vaswani2017attention} have shown their superiority in capturing long-term dependencies over LSTM based networks. Transformer based pretrained language models (PLM), such as GPT2 \cite{radford2019gpt2} and T5 \cite{JMLR:v21:20-074}, have been leveraged to build generative E2E TOD systems in the pretraining-and-finetuning paradigm, which have shown improved performance over LSTM-based ones. Examples include GPT2-based SimpleTOD \cite{hosseini2020simple}, SOLOIST \cite{peng2020etal}, AuGPT \cite{kulhanek2021augpt} and UBAR \cite{yang2021ubar}, and T5-based PPTOD \cite{su2021multitask} and MTTOD \cite{lee-2021-improving-end}, among others.
A drawback of existing PLM-based methods, viewed from efficiencies in memory, computation and learning, is that the whole history is used as the conditioning input at each turn. The dialog model thus becomes non-Markov across turns, i.e., the generation at current turn depends not only on the previous turn but also on all previous turns, namely the whole dialog history. 
Some models rely on all previous user utterances and system responses at each turn, and others like UBAR and MTTOD make the generation at current turn additionally conditional on all previous dialog states, DB results, system acts.
In both manners, the non-Markov models increase the memory and computation costs in both training and generation.
Furthermore, using the whole history increase model complexity and may hurt the training efficiency\footnote{In this paper, we use training efficiency to refer to a measure of quality of a statistical procedure \cite{everitt2010cambridge}, similar to sample efficiency.
Essentially, a more efficient procedure achieves lower error performance than a less efficient one under a given amount of samples.
}, especially when facing small amounts of labeled training data (the low-resource setting).

As briefly reviewed above, Markov models have been used in previous LSTM-based TOD systems \cite{wen2017a}-\cite{zhang-etal-2020-probabilistic}, but not studied for PLM-based TOD systems.
In this paper, motivated by the observation that dialog states could be viewed as Markov states \cite{williams2016dialog,sutton2018reinforcement}, we propose to build {M}arkovian {G}enerative {A}rchitectures (MGA) over PLM backbones for efficient TOD systems, which could be based on GPT2 or T5 but use shortened history \footnote{The code is released at \url{https://github.com/thu-spmi/MGA}}.
We follow the classic definition of the dialog state, which is defined to compactly summarize the dialog history from the beginning up to current turn \cite{williams2016dialog}.
It is also based on this definition that dialog states in widely-used dialog datasets such as MultiWOZ \cite{eric2019multiwoz} are annotated.
We observe that dialog states defined in this way essentially corresponds to Markov states from modeling the dialog as a Markov decision process \cite{sutton2018reinforcement}.
A shortened history, which consists of only last dialog state and system response along with current user utterance, can be supposed to provide enough context information for the agent to make prediction at current turn.
Theoretically, if the dialog state is strictly a Markov state, it means the future is independent of earlier history given the dialog state and there is no information lost in using the shortened history for predicting the future.

Remarkably, when comparing Markov and non-Markov models (both over PLM backbones) in practice, it is difficult to obtain general guarantees which one would be better.
For Markov models, assuming that the dialog state has the Markov property is a model assumption, and in practice, may be only approximately satisfied.
However, assuming that the non-Markov models, with increasing model complexity, can be well trained, under a given amount of samples, is also an assumption, and in practice, may be more easily over-fitted, particularly in the low-resource setting.
So regarding the comparison of Markov and non-Markov models (both over PLM backbones), there are some interesting research questions, which will be empirically addressed in this paper. 
First, in the rich-resource setting, can Markov models perform close to non-Markov models? This roughly tests the Markov model assumption.
Second, in the low-resource setting, can Markov models outperform non-Markov models? This compares the training efficiency of the two models.

Extensive experiments are taken on the widely used MultiWOZ2.1 dataset \cite{eric2019multiwoz}.
First, in the rich-resource setting, 100\% labeled data in MultiWOZ2.1 training set is used, and it is found that compared with existing non-Markov PLM-based systems, the proposed Markov PLM-based systems achieve equally strong performance, with significantly reduced computation and memory costs.
Second, we consider the low-resource setting, where some proportions (10\% and 20\%) of the labeled dialogs from MultiWOZ2.1 training set are drawn.
The rest dialogs in the training set are discarded for supervised-only training, or treated as unlabeled for semi-supervised training.
The proposed Markov models achieve much better than their non-Markov counterparts in both low-resource settings, which demonstrates the superiority of the Markov models in training efficiency.
Overall, these results give encouraging empirical evidences, which support the advantage of using Markovian architectures for PLM-based TOD systems.

\vspace{-0.8em}
\section{Background and Related Work}
\label{sec:related}
\vspace{-0.8em}
\textbf{~~~~Markov state}
A widely accepted theoretical model of task-oriented dialog systems is based on Markov Decision Processes (MDPs) to model the interaction between the agent (namely the dialog system) and the user (which is taken as the environment) \cite{young2013pomdp}.
As reviewed in \cite{williams2016dialog}, the dialog state is defined to summarize the dialog history up to current turn to a level of detail that provides sufficient information for choosing the next system action.
It can be seen that such concept of dialog state essentially corresponds to the Markov state, which is introduced in the context of MDP theory to handle partial observability of the environment state \cite{sutton2018reinforcement}.
Conceptually, the Markov state summarizes all relevant information in the trajectory history necessary for the agent to make decision.
This is exactly what the dialog state is defined to be.
See Chapter 17.3 in \cite{sutton2018reinforcement} for more expositions.

\textbf{LSTM-based Markov TOD systems}
The sequential turns of a dialog are usually modeled as a Markov process, not only in traditional modular approach but also in early E2E trainable approach.
At each turn, the dialog agent only reads the last dialog state, last system response and current user utterance, and generates the current dialog state and system response, without accessing the whole history.
Such Markov assumption underlies some recent E2E TOD systems \cite{lei2018sequicity,zhang2020task,zhang-etal-2020-probabilistic}.
Different encoder-decoder architectures are developed to model the conditional transitions.
For example, while SEQUICITY \cite{lei2018sequicity} uses two decoders to generate belief spans and responses respectively, DAMD \cite{zhang2020task} further introduces a third decoder to explicitly generate system acts. 
LABES \cite{zhang-etal-2020-probabilistic} extends SEQUICITY, as it introduces an additional encoder for dialog states and explicitly models the probabilistic transitions between dialog states across turns, and thus enables semi-supervised learning feasible.
Despite these differences, all these early E2E methods are LSTM-based.

\textbf{Non-Markov TOD systems}
Until very recently, non-Markov TOD systems have been developed, but with increasing model complexity \cite{hosseini2020simple}-\cite{lee-2021-improving-end}.
They heavily rely on the Transformer-based pretrained language models to provide good initialization for training large models.


\vspace{-0.5em}
\section{Method}
\vspace{-0.5em}
\label{sec:method}
In TOD systems, let $u_t$ denote the user utterance, $b_t$ the dialog state, $db_t$ the DB result, $a_t$ the system act and $r_t$ the delexicalized response, respectively, at turn $t=1,\cdots,T$, for a dialog of $T$ turns. In this work, all these variables are converted to token sequences, like in DAMD \cite{zhang2020task} (See examples in Fig.~\ref{fig:heat}).

\begin{figure}[t]
	\centering
	\subfigure[Architecture of SimpleTOD]
	{\label{fig:SimpleTOD}
	\includegraphics[width=0.9\columnwidth]{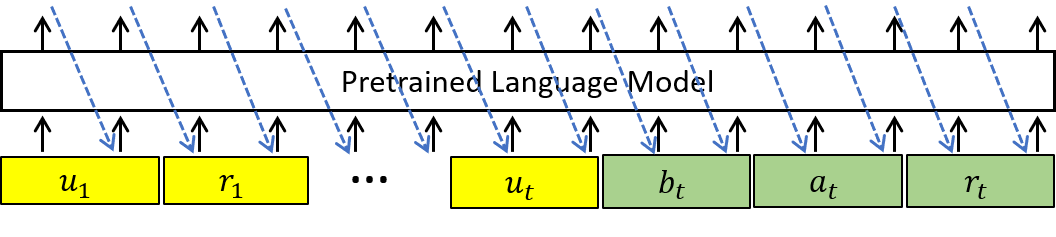} }
	\vspace{-0.5em}
	\subfigure[Architecture of UBAR]
	{\label{fig:UBAR}
	\includegraphics[width=0.9\columnwidth]{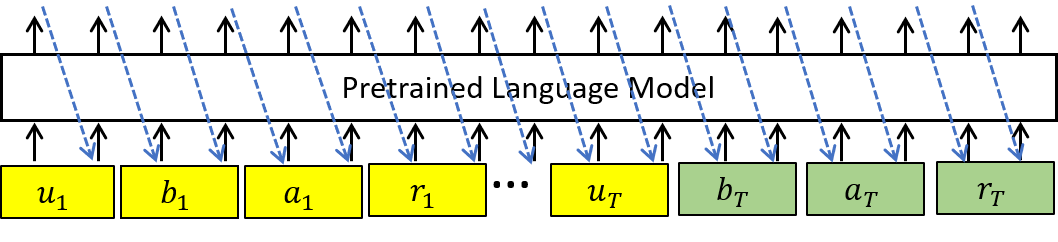} }
	\vspace{-0.5em}
	\subfigure[Architecture of MGA generative model]
	{\label{fig:MGA}
	\includegraphics[width=0.9\columnwidth]{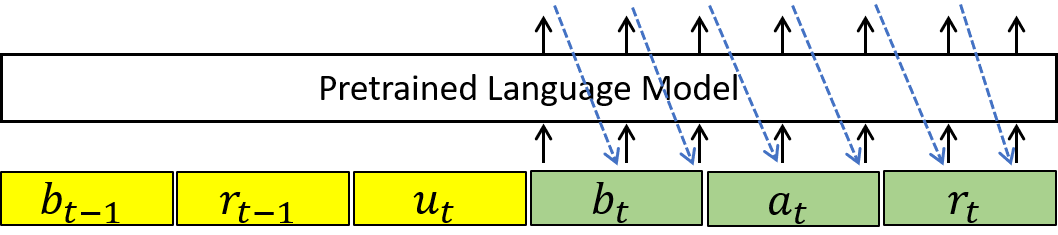}}
	\subfigure[Architecture of MGA inference model]
	{\label{fig:MGA-inf}
	\includegraphics[width=0.9\columnwidth]{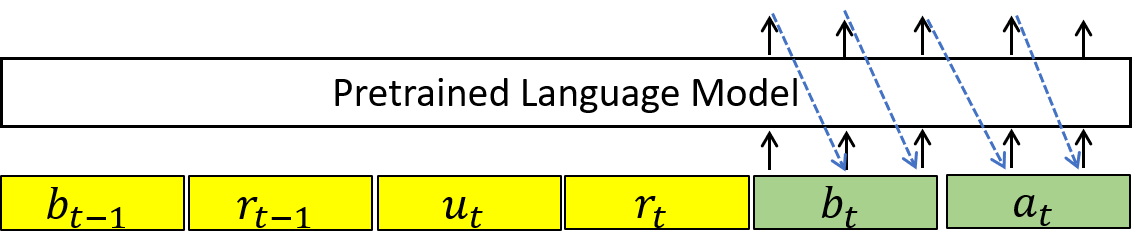} }
	\caption{Different architectures for $p(output|input)$ in different PLM-based TOD systems.
	Yellow boxes represent the conditioning $input$ of the model during generation, and green boxes the targeting $output$. 
}
	\label{fig:structure}
	\vspace{-0.5em}
\end{figure}

The work flow for a TOD system is, for each dialog turn $t$, to generate (or say, predict) $b_t$, $a_t$ and $r_t$ \footnote{Note that database result $db_t$ is deterministically obtained by querying database using the predicted $b_t$. We omit $db_t$ in the discussion for simplicity.}, given $u_t$ and dialog history $u_1, r_1, \cdots, u_{t-1}, r_{t-1}$. 
In recent PLM-based TOD systems, such work flow of a dialog system is unified into a single sequence generation problem, which can be accomplished by finetuning PLMs, such as GPT2 \cite{radford2019gpt2} and T5 \cite{JMLR:v21:20-074}\footnote{In this paper, when referring to PLMs, GPT2 and T5 are used as typical examples. GPT 2 is a decoder-only PLM and T5 is an encoder-decoder PLM. Both are based on self-attention based Transformer neural networks \cite{vaswani2017attention}.}.
The basic idea is that for building a particular conditional model, $p(output|input)$, where $input$ and $output$ are token sequences, a PLM can be finetuned over training samples $(input, output)$ (often referred to as training sequences \cite{hosseini2020simple}), and after finetuning, the model can be used for generation, i.e., generating $output$ after receiving $input$.
Different PLM-based dialog models employ different architectures for $p(output|input)$.
For example, the generative architecture used in SimpleTOD \cite{hosseini2020simple} is described by
\begin{equation} \label{eq:SimpleTOD}
p(b_t, a_t, r_t|u_1, r_1, \cdots, u_{t-1}, r_{t-1}, u_t),
\end{equation}
as shown in Fig. \ref{fig:SimpleTOD} and the one used in UBAR \cite{yang2021ubar} is
\begin{equation} \label{eq:UBAR}
p(b_t, a_t, r_t|u_1, b_1, a_1, r_1, \cdots, u_{t-1}, b_{t-1}, a_{t-1}, r_{t-1}, u_t),
\end{equation}
as shown in Fig. \ref{fig:UBAR}.


A drawback of existing PLM-based TOD systems is that they assume non-Markovian transitions across turns, as can be seen from the transition distributions in Eq. \eqref{eq:SimpleTOD} and \eqref{eq:UBAR}. 
This brings inefficiencies in memory and computation.
Furthermore, using the whole history increase model complexity and may hurt the training efficiency, especially when facing small amounts of labeled training data (the low-resource setting).

To address this drawback, we propose to build Markovian generative architecture (MGA) over PLM backbones, which, with parameter $\theta$, is designed to be
\begin{equation} \label{eq:MGA}
p_{\theta}(b_t, a_t, r_t|b_{t-1}, r_{t-1}, u_t),
\end{equation}
as shown in Fig. \ref{fig:MGA}.
Specifically, to predict $b_t$, $a_t$ and $r_t$ at each turn $t$, we propose to use only the dialog state $b_{t-1}$ and response $r_{t-1}$ from previous turn along with current user utterance $u_t$, instead of using the whole dialog history, as the conditioning input.
This is motivated by the observation that dialog states could be viewed as Markov states \cite{williams2016dialog,sutton2018reinforcement}, as explained in Section \ref{sec:related}.
Dialog state $b_{t-1}$ is defined (and annotated) to be a compact summary of dialog history up to turn $t-1$, which, together with $r_{t-1}$ and $u_{t}$, can be supposed to carry enough context information for the agent to make prediction for $b_t$, $a_t$ and $r_t$.



In the following, this paper provides encouraging empirical evidences that support the Markov model assumption, by evaluating Markov TOD systems in both rich-resource and low-resource settings and comparing to their non-Markov TOD systems.
In supervised learning, the generative model in Eq. \eqref{eq:MGA} can be finetuned from PLMs such as GPT2 or T5, by maximizing the following conditional likelihood:




\begin{equation}
\label{eq:sup-ds}
\begin{aligned}
    \mathcal{J}_{\text{MGA}} &=\log p_{\theta}(b_t, a_t, r_t|b_{t-1}, r_{t-1}, u_t)\\
    &= \sum_{i=1}^{|b_t \oplus a_t \oplus r_t|} \log p_\theta(c_i| b_{t-1}, r_{t-1}, u_t, c_{<i}) 
\end{aligned}
\end{equation}
where $\oplus$ denotes the concatenation of sequences, $|b_t \oplus a_t \oplus r_t|$ denotes the length of $b_t \oplus a_t \oplus r_t$ in tokens and $c_i$ denotes the $i$-th token in  $b_t \oplus a_t \oplus r_t$.

\begin{table*}[t]
\centering
\resizebox{0.8\linewidth}{!}{
\begin{tabular}{lcccccc}
\hline
\textbf{Model} &Backbone &Markovian &Inform &Success &BLEU &Combined\\
\hline
DAMD$^{\dagger}$ \cite{zhang2020task} & LSTM &Y &57.9 &47.6 &16.4 &69.2\\
LABES$^{\dagger}$ \cite{zhang-etal-2020-probabilistic} & LSTM &Y &68.5 &58.1 &18.9 &82.2\\
UBAR$^{\dagger}$ \cite{yang2021ubar} &GPT2 &N &83.4	&70.3 &17.6	&94.4\\
MTTOD$^{\dagger}$ \cite{lee-2021-improving-end} &T5 & N &85.4 &75.7 &19.64 &100.2\\
\hdashline
SimpleTOD \cite{hosseini2020simple} &GPT2 &N &82.23$\pm$1.09 &69.13$\pm$0.65 &18.23$\pm$0.26 &93.92$\pm$1.11\\
UBAR \cite{yang2021ubar} &GPT2 & N &84.60$\pm$1.00 &73.13$\pm$1.03 &18.54$\pm$0.11 &97.40$\pm$1.03\\
MGA-GPT2 &GPT2 &Y &84.50$\pm$0.29  &72.77$\pm$0.50  &18.96$\pm$0.36  &97.59$\pm$0.54\\
MTTOD \cite{lee-2021-improving-end} &T5 &N &85.67$\pm$0.61 &75.97$\pm$0.45 &19.62$\pm$0.23 &100.43$\pm$0.67\\
MGA-T5 &T5 &Y &84.63$\pm$0.82  &75.17$\pm$1.11  &19.49$\pm$0.15  &99.39$\pm$0.63\\
\hline
\end{tabular}
}
\caption{Results on MultiWOZ2.1. Above the dashed line are cited from the official website of MultiWOZ (marked with ${\dagger}$), and below are statistical results obtained by our own runs with 3 different random seeds.}
\label{tab:baseline}
\end{table*}

\begin{table}[t]
\centering
\resizebox{0.8\linewidth}{!}{
\begin{tabular}{lccc}
\toprule
Model &Inform &Success &BLEU\\
\midrule
MGA-GPT2 vs SimpleTOD &0.213 &0.006 &0.001 \\
MGA-GPT2 vs UBAR &0.116 &0.863 &0.067\\
MGA-T5 vs MTTOD &0.110 &0.290 &0.494\\
\bottomrule
\end{tabular}
}
\caption{P-values for comparisons between different models.}
\label{tab:p-value}
\end{table}

\begin{table}[t]
\centering
\resizebox{0.7\linewidth}{!}{
\begin{tabular}{lcc}
\toprule
Model & Training Time & Generation Time over Test Set\\
\midrule
SimpleTOD &396 min &276 s\\
UBAR &369 min &352 s\\
MGA-GPT2 &204 min &229 s\\
\hdashline
MTTOD  &316 min & 756 s\\
MGA-T5 &235 min & 734 s\\
\bottomrule
\end{tabular}
}
\caption{Time costs of different models in training and testing.}
\vspace{-0.6em}
\label{tab:speed}
\end{table}

In the low-resource setting, we further consider semi-supervised learning in addition to supervised-only learning.
For running variational semi-supervised learning with the generative model in Eq. \eqref{eq:MGA}, an inference model also with Markovian architecture can be similarly designed as follows:
\begin{equation} \label{eq:MGA-inf}
q_{\phi}(b_t, a_t|b_{t-1}, r_{t-1}, u_t,r_t),
\end{equation}
as shown in Fig. \ref{fig:MGA-inf}.
The supervised training of Eq. \eqref{eq:MGA-inf} and the variational learning of Eq. \eqref{eq:MGA} and \eqref{eq:MGA-inf} are omitted to save space, which are similar to the non-Markovian models in \cite{liu2021variational}.

\section{Experiments}
\label{sec:exp}
\subsection{Datasets}
Experiments are conducted on MultiWOZ2.1 \cite{eric2019multiwoz}. MultiWOZ2.1 is an English multi-domain TOD dataset, collected via human-to-human interactions using the Wizard-of-Oz approach. It consists of a total of 10.4k dialog sessions, spanning over seven domains (restaurant, train, attraction, hotel, taxi, hospital, police). Each dialog contains 6.85 turns on average.
Compared to MultiWOZ2.0, MultiWOZ2.1 removed some noisy state values.

\vspace{-0.5em}
\subsection{Evaluation metrics}
Different dialog systems are evaluated in the end-to-end setting, which means that the generated dialog states and system acts are used in response generation.
To avoid any inconsistencies in evaluation and rigorously compare our models with others, we use the standardized evaluation scripts in \cite{nekvinda-dusek-2021-shades}, which are now also the scripts adopted in the MultiWOZ website.
There are mainly four metrics for corpus-based evaluation. \emph{Inform Rate} measures how often the entities provided by the system are correct. \emph{Success Rate} refers to how often the system is able to answer all the requested attributes by user. \emph{BLEU Score} is used to measure the fluency of the generated responses. And the \emph{Combined Score} is computed as (BLEU + 0.5 * (Inform + Success)).


\begin{figure}[t]
\centering
\includegraphics[width=0.85\linewidth]{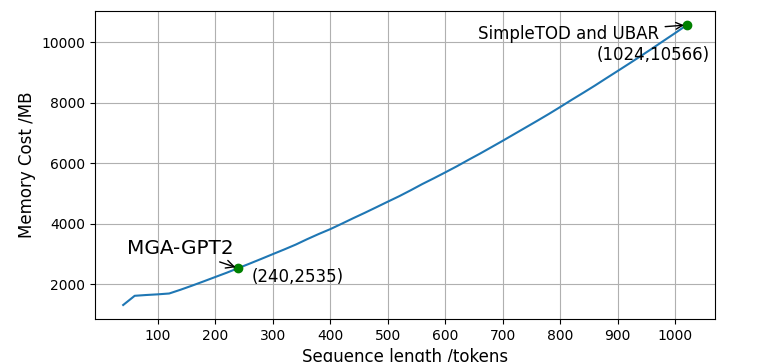}
\caption{The memory costs during training with batch size 4, as a function of the lengths of training sequences.
For MGA-GPT2, SimpleTOD and UBAR, the means and standard deviations of the lengths of training sequences are 98$\pm$30, 190$\pm$112 and 440$\pm$220, respectively. The maximum sequence lengths for the three models are marked in the figure.}
\label{fig:memory}
\vspace{-0.8em}
\end{figure}

\vspace{-0.5em}
\subsection{Results in the rich-resource setting}
In the rich-resource setting, 100\% labeled data in MultiWOZ2.1 training set are all used for supervised training of different TOD models. 
The results are summarized in Table~\ref{tab:baseline}.
We experiment with both Markov and non-Markov models over different PLM backbones (GPT2 and T5).
We use distilled GPT2 (82M) and T5-base (220M).
The MGA models trained with the backbones of GPT2 and T5 are denoted by MGA-GPT2 and MGA-T5 respectively.
By a plain reading of mean Combined Scores, it can be seen that MGA-GPT2 achieves the best performance among all the GPT2 based models and MGA-T5 performs slightly worse than MTTOD. However, the differences are not large.

In order to rigorously compare different models, we conduct significance tests between the proposed MGA models and the state-of-the-art (SOTA) models with different backbones of GPT2 and T5, respectively. The McNemar test is used for \emph{Inform} and \emph{Success} and the matched pair test for \emph{BLEU} \cite{gillick1989some}. The p-values are reported in Table~\ref{tab:p-value}. 
When comparing MGA-GPT2 with SimpleTOD, the p-values for Success and BLEU are less than 0.01, which shows that the improvement of MGA-GPT2 over SimpleTOD is somewhat significant.
All other p-values are pretty large ($>$0.05), indicating that the differences between the models being compared are in fact not significant.
Therefore, MGA-GPT2 in fact performs close to UBAR, and MGA-T5 is close to MTTOD.
Note that in the rich-resource setting, non-Markov models can be well trained. The comparable performances between Markov and non-Markov models in the rich-resource setting could be viewed as supporting evidences for the Markov assumption that dialog states as used in our experiments could be viewed as Markov states and there is almost no information lost in using the shortened history for predicting the future. 
Encouragingly, compared with non-Markov models, Markov models achieve equally strong results and at the same time, as detailed below, can significantly reduce memory and time costs.

For memory cost, it is obvious that the training sequences for Markov models are generally much shorter than non-Markov models. For example, it can be seen from Fig.~\ref{fig:memory} that MGA-GPT2 consumes much less memory (reduced by 76\%) than SimpleTOD and UBAR during training.
The memory costs for MTTOD and MGA-T5 are 22,981MB and 12,641MB respectively during training with batch size 8.
For time cost, we run different models on a single GPU (GPT2 based models over Tesla P100 and T5 based models over Tesla V100), record the total training time and the inference time on test set and report them in Table~\ref{tab:speed}. The same training strategy is used to train all models, where the maximum epoch is 50 and early stopping is applied. It is clear from Table \ref{tab:speed} that Markov models significantly reduces training time and inference time, compared to the their non-Markov counterparts.

\begin{table}[t]
	\centering
	\resizebox{\linewidth}{!}{
			\begin{tabular}{cl cccc}
				\toprule
				\multicolumn{2}{c}{Model Configuration} &\multicolumn{4}{c}{MultiWOZ2.1}\\ 
				\cmidrule(lr){1-2} \cmidrule(lr){3-6} 
				Label Proportion &Method  &Inform &Success &BLEU &Combined\\ 
				\midrule
				\multirow{4}{*}{20\%} &Supervised Non-Markov  &65.30 & 50.70 & 16.79 & 74.79\\
				&Supervised Markov  &75.25  &59.90  &16.73  &{84.31} \\
				&Semi-Supervised Non-Markov &72.40 & 60.30 & 18.00 & 84.35\\
				&Semi-Supervised Markov  & 80.90  & 67.40  & 18.53  &{92.68}\\ 
				\midrule
				\multirow{4}{*}{10\%} &Supervised Non-Markov   &48.20 & 35.70 & 15.00 & 56.95 \\
				&Supervised Markov  &67.75  &49.60  &15.36  &{74.03}\\
				&Semi-Supervised Non-Markov   &69.60 & 58.30 & 16.69 & 80.64\\ 
				&Semi-Supervised Markov   &79.75  &66.10   & 16.78  &{89.89}\\
				\midrule
		\end{tabular}}
		\caption{Results for supervised-only and semi-supervised learning of different GPT2 based models.}
		\label{tab:semi-results}
		\vspace{-1em}
\end{table}

\vspace{-0.5em}
\subsection{Results in the low-resource setting}
In the rich-resource setting, the labeled data are rich so that the effects of training efficiency of different models on the performances diminish.
To further evaluate the Markov and non-Markov models in training efficiency, we randomly draw some small proportions of labeled training data (10\% and 20\%) from MultiWOZ2.1.
The rest dialogs in the training set are discarded for supervised-only training, or treated as unlabeled for semi-supervised training.
In the task of supervised-only training, only the small amounts of labeled data are used to train the models by supervised learning, which are then used for testing.
In the task of semi-supervised training, after trained with the small amounts of labeled data, the models are further trained using both labeled and unlabeled data by variational learning, as described in \cite{liu2021variational}.
The results from the two training tasks are shown in Table \ref{tab:semi-results}, where the Markov model is MGA-GPT2, the non-Markov model follows the generative model in \cite{liu2021variational}, and both use GPT2 as the backbone.
It can be seen from Table \ref{tab:semi-results} that the Markov model outperforms the non-Markov model by large margins consistently across all conditions (with different labeling proportions and in both supervised-only and semi-supervised tasks). These results clearly show the advantage of the Markov model over the non-Markov models in training efficiency, and the gains are more significant in the limited training data scenarios.

\vspace{-0.5em}
\subsection{Analysis}
\label{sec:analysis}

\begin{figure}[t]
	\centering
	\begin{minipage}[t]{0.49\linewidth}
	\subfigure[UBAR]
	{\label{fig:attention_ubar}
	\includegraphics[width=\linewidth]{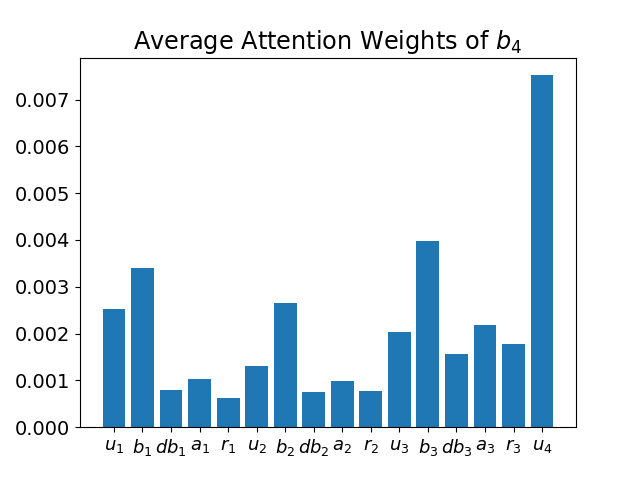}}
    \end{minipage}
    \vspace{-0.5em}
    \begin{minipage}[t]{0.49\linewidth}
	\subfigure[MGA-GPT-2]
	{\label{fig:attention_mga}
	\includegraphics[width=\linewidth]{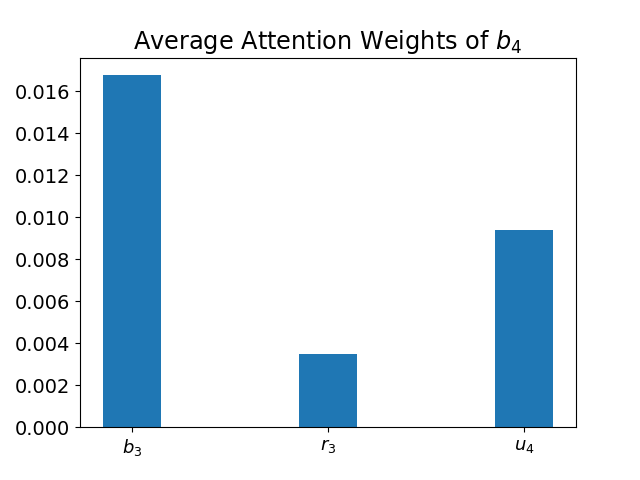} }
    \end{minipage}
    \vspace{-0.5em}
	\caption{Average attention weights of UBAR and MGA-GPT2 for predicting the dialog state in the 4-th turn.}
	\label{fig:attention}
	\vspace{-0.5em}
\end{figure}


In building TOD systems, dialog states are defined and annotated to compactly summarize the dialog history from the beginning up to current turn. 
In this sense, dialog states essentially correspond to Markov states, which contain sufficient information for the agent to make prediction and decision. 
Thus, $b_{t-1}$, $r_{t-1}$ and $u_t$ could be sufficient for the agent to generate $b_t$, $a_t$ and $r_t$. This is the hypothesis in this paper that motivates us to develop Markovian generative architectures over PLM backbones.
In addition to the previous experiments to validate this hypothesis, we further provide the following analysis to help readers to appreciate this hypothesis.

\begin{figure*}[t]
    \subfigure{\includegraphics[width=\linewidth,height=00.26\linewidth]{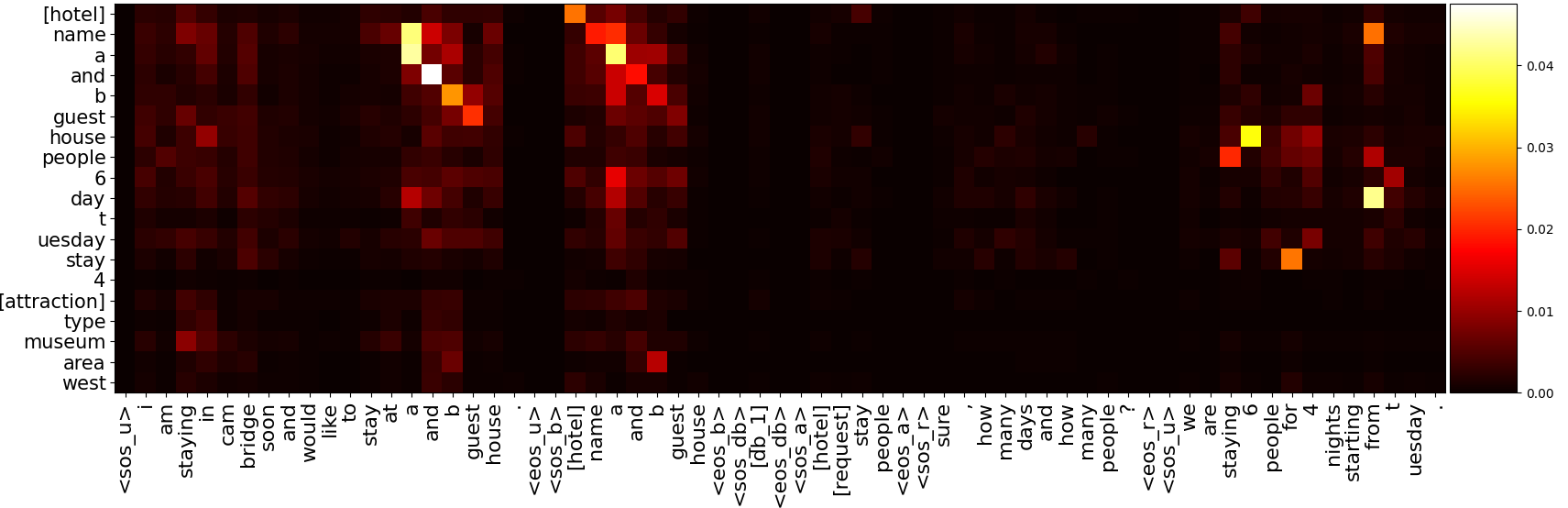}}
    \subfigure{\includegraphics[width=\linewidth,height=00.26\linewidth]{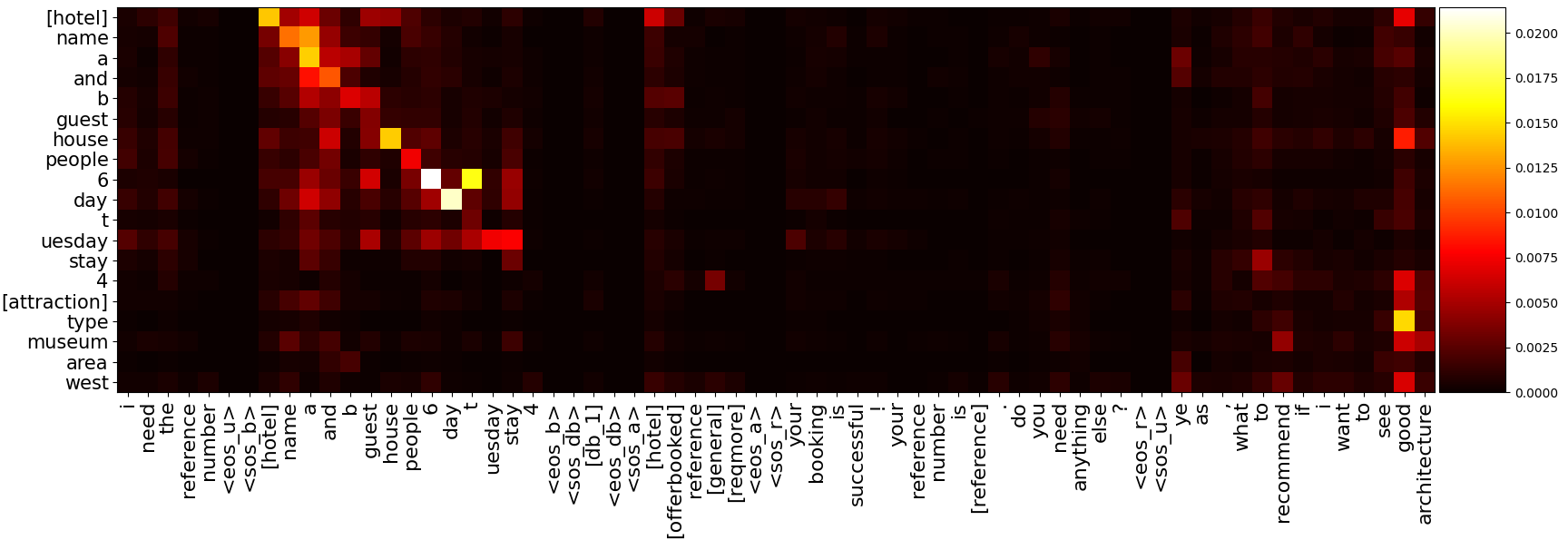}}
    \subfigure{\includegraphics[width=\linewidth,height=00.26\linewidth]{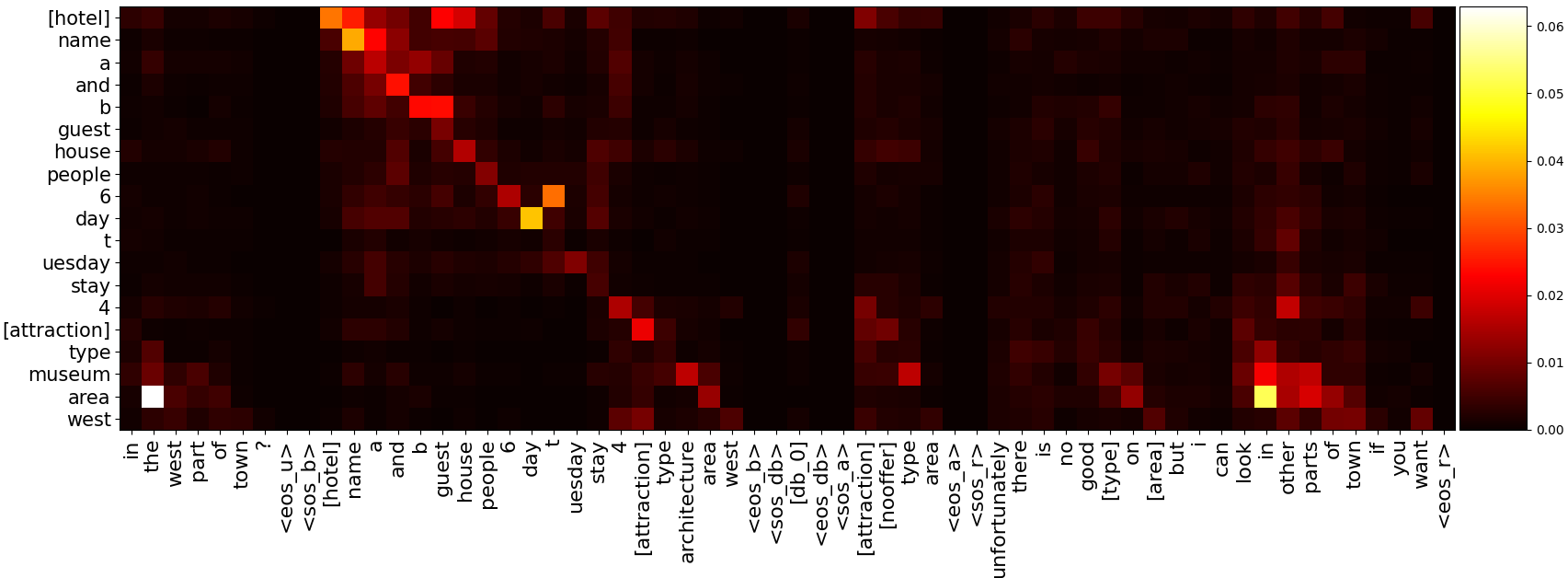}}
	\caption{The attention heatmap of UBAR. The vertical axis represents the dialog state at the $4$-th turn (i.e., $b_4$). The horizontal axis represents the user utterances $u_t$, belief states $b_t$, db results $db_t$, system acts $a_t$ and system responses $r_t$ from all previous turns, which corresponds to the text spans between $<sos\_*>$ and $<eos\_*>$ respectively, where $* = u, b, db, a, r$.
	}
	\label{fig:heat}
\vspace{-0.5em}
\end{figure*}

We calculate the average attention weights, which represent how much the model attends to previous turns when they make predictions in a certain turn $t$.
We compare MGA-GPT2 with UBAR \cite{yang2021ubar}. Consider $t=4$ and Fig.~\ref{fig:attention} shows the average attention weights over test set during generating $b_4$.
From Fig.~\ref{fig:attention_ubar}, we can see that in UBAR, dialog state $b_4$ mainly attends to current user utterance ($u_4$) and dialog states of all previous turns ($b_1$, $b_2$ and $b_3$). 
Note that dialog state is defined as the accumulation of history information, which means that $b_3$ actually contains almost all the slots and values of $b_1$ and $b_2$. Thus, the attentions to $b_1$ and $b_2$ are not essential, they appear mainly because there are no mechanisms to constrain such attentions in UBAR. 
In Fig.~\ref{fig:heat}, we provide an example to show how UBAR attends to history information when generating $b_4$.
It can be seen that the tokens in $b_1$ and $b_2$ with major attentions are all contained in $b_3$. 
This suggests that, if we do not let the model attend to $b_1$ and $b_2$, the model will naturally attend more to $b_3$ and still not miss the information in $b_1$ and $b_2$. 
This can be indeed observed from the attention weights of the proposed MGA-GPT2, as shown in Fig.~\ref{fig:attention_mga}. We can see that MGA-GPT2 attends much more to $b_3$ when generating $b_4$, while in UBAR, these attentions are scattered across $b_1,b_2,b_3$.
\section{Conclusion and Future Work}
A drawback of existing PLM-based models is their non-Markovian architectures across turns, which brings inefficiencies in memory, computation and learning.
Note that dialog states are defined and annotated to compactly summarize the dialog history, and essentially correspond to Markov states. 
In this paper, we propose to build Markovian Generative Architectures (MGA) over PLM backbones for efficient TOD systems, which could, but not limited to, be based on GPT2 and T5.
Experiments are conducted on MultiWOZ2.1 in both rich-resource and low-resource settings.
The efficiency advantages of the MGA models in reducing memory and computation are obvious.
In the rich-resource setting, the performances of the MGA models are on par with the SOTA non-Markov models (UBAR and MTTOD) under rigorous significance tests with different PLM backbones (GPT2 and T5).
The training efficiency of the MGA models is more evident in the low-resource settings for both supervised-only and semi-supervised tasks.
Developing Markov dialog models may have more important implications for reinforcement learning of the TOD systems. As remarked in \cite{sutton2018reinforcement}, ``We want our states to be compact as well as Markov'' (p384). A further study of MGA for sample efficient reinforcement learning would be very interesting future work.

\bibliographystyle{IEEEbib}
\bibliography{MGA}

\end{document}